%% file: main.tex
\documentclass[10pt,twocolumn,letterpaper]{article}

\usepackage{wacv}              
\usepackage{xr}
\usepackage{graphicx}


\definecolor{wacvblue}{rgb}{0.21,0.49,0.74}
\usepackage[pagebackref,breaklinks,colorlinks,allcolors=wacvblue]{hyperref}

\def\wacvPaperID{613} 
\def\confName{WACV}
\def\confYear{2027}

\title{IncR-3DSGG: Incremental Posed RGB Image Based 3D Scene Graph Generation via Privileged Depth Supervision}

\author{
  Mrunmai Phatak$^{1}$ \qquad Daniel Kienzle$^{1}$ \qquad Roman Küble$^{2}$ \qquad Jörg Hähner$^{2}$ \qquad Rainer Lienhart$^{1}$\\[0.3em]
  $^{1}$Machine Learning and Computer Vision Lab \qquad $^{2}$Organic Computing Group\\
  Universität Augsburg, Germany\\
  {\tt\small \{mrunmai.phatak, daniel.kienzle, roman.kueble, joerg.haehner, rainer.lienhart\}@uni-a.de}
}

\begin{document}
\maketitle
\input{sec/0_abstract}    
\input{sec/1_intro}
\input{sec/2_formatting}
\input{sec/3_finalcopy}
\input{sec/04_results_cam_ready}
\input{sec/5_conclusion}
{
    \small
    \bibliographystyle{ieeenat_fullname}
    \bibliography{main}
}

\end{document}

%% file: sec/0_abstract.tex
\begin{abstract}
Scene graphs provide a compact, structured representation of environments by
capturing objects and their spatial or semantic relationships in a form suitable
for embodied agents. Existing 3D scene graph generation methods often rely on
depth sensing at inference, either to associate object instances across views via
point cloud overlap or to compute inter-object relations from geometric cues,
limiting deployment to platforms with depth sensors. We propose IncR-3DSGG, which
builds 3D scene graphs incrementally from posed\footnote{RGB images with camera
extrinsics estimated by an off-the-shelf visual SLAM system.} RGB images,
eliminating the use of depth sensors at inference. Our method leverages a
Parametric Relation Predictor (PRP) that learns to estimate pairwise distances and
relative angles between objects from 2D visual features and 2D geometric cues,
using 3D geometric supervision available only during training. These learned
pairwise measurements are fused across views through Pose Graph
Optimization (PGO)~\cite{grisetti2010tutorial} to recover metrically grounded 3D
object positions and build a globally consistent 3D scene graph incrementally. 
On the real-world 3RScan~\cite{wald2019rio} dataset, our method attains $52.2$
mean object recall and $30.8$ mean predicate recall under the
3DSSG~\cite{wald2020learning} evaluation protocol, against $27.0$ and $17.6$ for
MonoSSG~\cite{wu2023incremental}, while using no depth or reconstructed geometry
at inference.
This shows that relations and metrically grounded 3D object
representations can be predicted from posed RGB images alone.
\end{abstract}

%% file: sec/1_intro.tex
\section{Introduction}
\label{sec:intro}

\begin{figure}[t]
    \centering
    \includegraphics[width=\columnwidth]{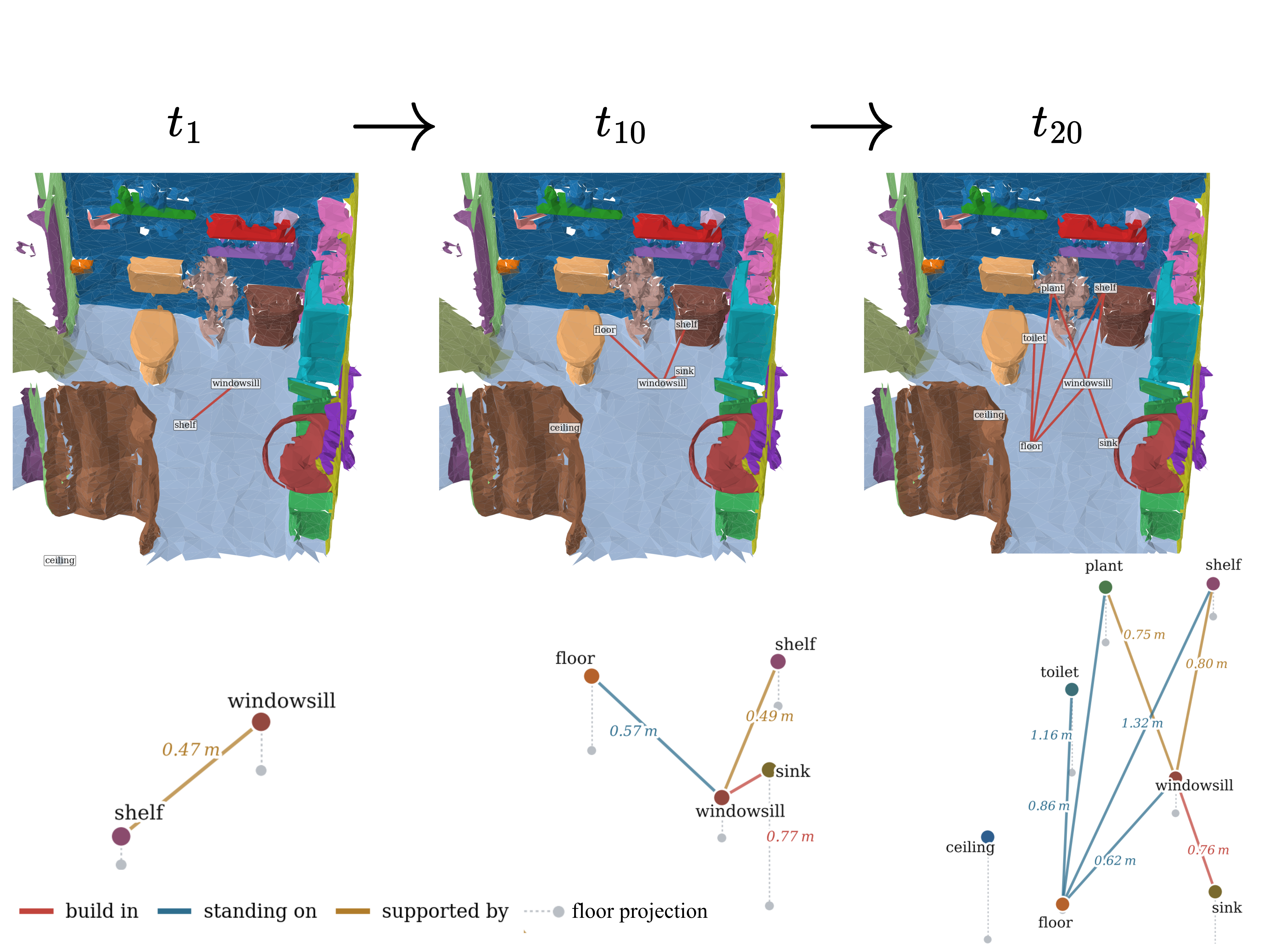}
    \caption{Incremental construction of the global scene graph $\mathcal{G}_G$ over $20$
processed views. Nodes are instantiated when an object is first detected and
refined as later views contribute pairwise constraints. The scan is drawn in the
global coordinate frame purely as a backdrop, showing how our predicted nodes
align with the annotated instances. Our method observes no depth and builds no
point cloud or mesh, producing object positions and relations directly.}
\label{fig:teaser}
\end{figure}

Scene understanding is a crucial area in computer vision for various tasks 
like image captioning, visual question answering and embodied 
reasoning~\cite{hildebrandt2020scene, singh2023scene, gu2024conceptgraphs}. 
An embodied agent navigating in unknown environment must build and maintain 
a map of its surrounding to support reasoning, planning and interactions. 
Conventional approaches for environment representation rely on SLAM-based 
methods~\cite{cadena2017past}, which construct geometric or topological maps to 
support agent localization and path planning. 
While this approach is effective for metric navigation, such representations 
lack object-level and semantic-level understanding of the environment as they only 
capture where the agent is, but not what surrounds it or how entities relate to 
one another.

Scene graphs provide compact yet expressive 
formalism~\cite{johnson2015image, chang2021comprehensive, lorenz2025copa}, 
by encoding objects, their attributes and inter-object 
relationships as a structured graph. Unlike purely metric maps, scene graphs can capture 
hierarchical information spanning object instances, rooms 
and spatial relations, enabling higher-level reasoning for 
tasks such as object goal navigation, language-grounded 
planning and manipulation~\cite{hughes2022hydra, werby2024hierarchical, gu2024conceptgraphs}.
Compared to dense 3D point cloud representations, 
scene graphs are a substantially more memory-efficient 
format for storing environmental information while preserving 
the semantic and relational content necessary for high-level 
reasoning. This makes them well suited to embodied settings 
where perceptual data must be processed and stored 
incrementally under tight computational budget.

Scene graph research is broadly divided into two 
paradigms: 2D scene graph generation which extracts 
object relationships from individual images~\cite{xu2017scene, tang2019learning, yang2018graph} 
and 3D scene graph generation which constructs scene 
graphs from 3D reconstructed points clouds or RGB-D 
sequences~\cite{wald2020learning, wu2021scenegraphfusion, gu2024conceptgraphs}. 
While 2D scene graph methods have made significant 
progress, they operate within the image plane. Hence, their spatial relationships remain viewpoint-dependent and inconsistent 
across camera perspectives~\cite{chang2021comprehensive}, lacking the geometric 
grounding that navigation and manipulation require. 3D scene graphs address this by  
capturing view-consistent spatial/semantic relationships 
anchored in the physical world~\cite{hughes2022hydra, rosinol2021kimera}.

Most prior work on 3D scene graphs assumes access to a complete 3D reconstruction or 
dense RGB-D sequences available a priori. The problem 
of incrementally building a scene graph as the agent 
actively explores has received comparatively limited 
attention, despite being the relevant setting for 
embodied AI. Moreover, methods that do operate incrementally 
such as SceneGraphFusion~\cite{wu2021scenegraphfusion} 
and ConceptGraphs~\cite{gu2024conceptgraphs}, rely on 
depth sensing at every frame to associate object instances 
across views via 3D point cloud overlap and to compute 
inter-object relations from geometric cues. This restricts deployment to platforms 
with depth sensors and adds computational overhead for point cloud maintenance and 
matching. MonoSSG~\cite{wu2023incremental} removes the depth sensor, however reads object
geometry off a sparse ORB-SLAM3~\cite{campos2021orb} keypoint map, so an
object is representable only if enough of its points triangulate.

In this work, we address the problem of incrementally constructing a metrically grounded 3D 
scene graph from a sequence of posed RGB images, without depth sensing at 
inference. 
Our key insight is that, while 3D geometric reasoning 
is essential for accurate scene graph construction, the 3D 
signal itself is not explicitly required at the 
deployment. Instead, we train a Parametric Relation Predictor (PRP) 
with privileged\footnote{We use the attribute \emph{privileged} to 
express that 3D supervision (depth, point clouds) is 
available only at training time, but not at inference.}
 depth supervision, to estimate pairwise 3D spatial geometry 
 including metric distances, angular displacements, and contact-indicative 
 features directly from 2D visual observations and panoptic segmentation 
 features. At inference these predictions serve as noisy but 
informative measurements in a pose graph optimization (PGO) framework, which 
jointly recovers globally consistent object positions across an incrementally 
growing set of views.

\begin{figure*}[!ht]
    \centering
    \includegraphics[width=0.7\textwidth]{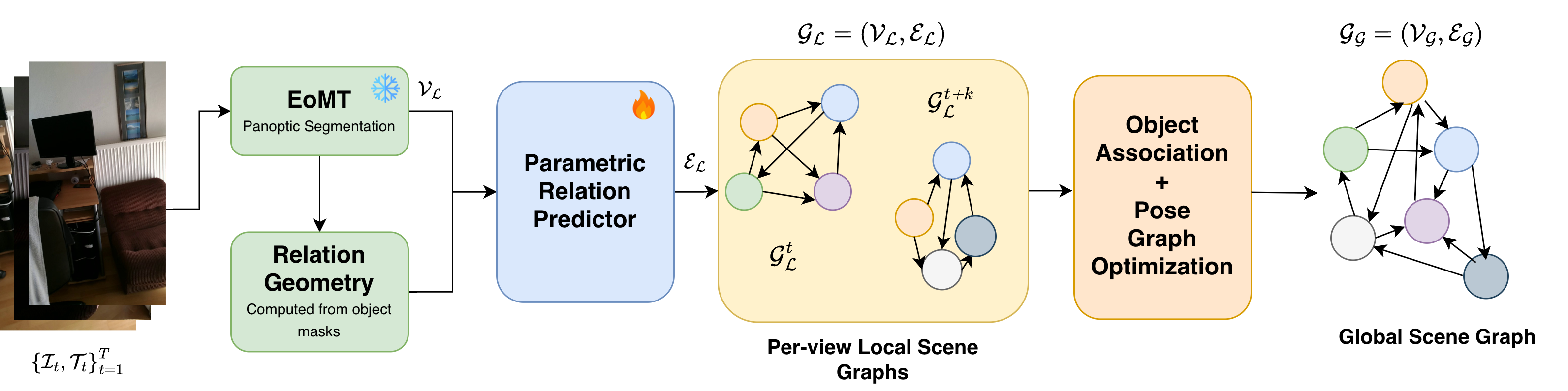}
    \caption{Overview of our IncR-3DSGG framework. Given an RGB image $\mathcal{I}_t$ and camera poses $\mathcal{T}_{t=1}^T$, a frozen EoMT backbone produces per-object nodes $\mathcal{V}_\mathcal{L}$ and 2D relation geometry, which are converted into edges $\mathcal{E}_\mathcal{L}$ using learned Parametric Relation Predictor, forming per-view local scene graphs $\mathcal{G}_\mathcal{L}$. These are incrementally merged via cross-view object association and pose graph optimization into a globally consistent 3D scene graph $\mathcal{G}_\mathcal{G}$.
}
\label{fig:piepile_incr}
\end{figure*}

Unlike prior approaches that localize each object independently from
depth-projected point clouds, we propagate spatial information through the
relational constraint graph, so well-observed landmarks improve the localization
of rarely seen objects. Figure~\ref{fig:teaser} shows a qualitative example of a
3D scene graph built incrementally by our method as views are processed.
Figure~\ref{fig:piepile_incr} shows the full pipeline, which takes only
posed RGB
images and produces a metrically grounded 3D scene graph.
The contributions are:

\begin{itemize}
\item We present IncR-3DSGG, a framework for incremental 3D scene graph generation
from posed RGB images, removing depth dependence at inference through privileged
3D supervision during training.
\item We introduce a Parametric Relation Predictor (PRP) that estimates pairwise
metric distances, angular displacements, and contact features from 2D input alone,
learning metric constraints without explicit 3D reconstruction.
\item We propose a pose graph optimization (PGO) formulation that recovers globally
consistent 3D object positions from the predicted parametric relations.
\item On the real-world 3RScan dataset we attain $52.2$ mean object recall and $30.8$
mean predicate recall under the 3DSSG evaluation protocol, against $27.0$ and
$17.6$ for MonoSSG~\cite{wu2023incremental}, while using no depth or
reconstructed geometry at inference.
\end{itemize}

\noindent Code and model checkpoints are published at \url{https://mrunmaivp.github.io/IncR-3DSGG/}.

%% file: sec/2_formatting.tex
\section{Related Work}
\label{sec:related_work}

\subsection{Spatial Mapping and SLAM}
Simultaneous Localization and Mapping (SLAM)~\cite{smith1990estimating} enables an
agent to build a metric map of an unknown environment while tracking its own pose
within it. Classical visual SLAM splits into feature-based
methods~\cite{mur2015orb}, which match keypoint descriptors across frames, and
direct methods~\cite{engel2014lsd, engel2017direct}, which minimize photometric
error. More recently, feed-forward models such as VGGT~\cite{wang2025vggt} regress
camera parameters, depth and point maps in a single pass, though only up to an
unknown global scale. All produce geometrically accurate maps, but treat the
environment as spatial coordinates without semantic or relational structure,
answering \textit{where am I} rather than \textit{what is around me and how is it
arranged}.

Semantic SLAM approaches such as Kimera~\cite{rosinol2021kimera} fuse per-frame
segmentation with reconstruction, labelling each point with a class. This supports
queries like ``where are the chairs,'' but operates on categories rather than
instances and captures no relations between objects. Scene graphs address both
limitations, encoding object instances, attributes, and pairwise relations as a
structured graph.

\subsection{Scene Graph Generation}
Unlike the metric maps produced by SLAM, scene graphs~\cite{johnson2015image,
xu2017scene} capture scene composition as structured graphs whose nodes are object
instances and edges encode spatial or semantic relationships. This relational
structure directly supports task planning~\cite{agia2022taskography}, object
search~\cite{yang2018visual}, and visual question answering~\cite{teney2017graph},
which are difficult with unstructured point clouds or occupancy grids.

Early 3D scene graph generation operates on pre-existing reconstructions of the
environment. 3DSSG~\cite{wald2020learning} takes segmented point clouds from complete room
scans and predicts pairwise relations between object instances with a graph neural
network (GNN) over point cloud geometry. Such methods assume a fully reconstructed and
segmented 3D scene, making them unsuitable for incremental operation where the
environment is observed progressively through partial views.

Recent methods construct scene graphs incrementally as observations arrive.
SceneGraphFusion~\cite{wu2021scenegraphfusion} maintains a volumetric object map
updated with each RGB-D observations through category-gated proximity matching, 
predicting relations with a GNN over the evolving
map. ConceptGraphs~\cite{gu2024conceptgraphs} fuses per-view detections into an
open-vocabulary global map, associating objects by geometric overlap and
CLIP~\cite{radford2021learning} feature similarity, and taking object positions as
the centroid of accumulated depth-projected points. Hydra~\cite{hughes2022hydra}
builds hierarchical scene graphs in real time over a dense metric-semantic mesh,
and HOV-SG~\cite{werby2024hierarchical} extends the hierarchical formulation with
open-vocabulary semantics from vision-language models. Closest to our setting, 
MonoSSG~\cite{wu2023incremental} 
removes the depth sensor 
by reconstructing a sparse point map from RGB sequence using ORB-SLAM3~\cite{campos2021orb} and estimating scene graphs over the 
resulting entities. Object geometry is nonetheless read from the reconstructed 3D 
points, hence the representation degrades wherever few points are triangulated.

A common requirement across these methods is that object positions come from
back-projected depth or point maps, either as centroids of accumulated point clouds or running
averages of depth-derived world coordinates. This restricts deployment to
platforms with active depth sensors and leaves the map sensitive to depth noise
and sensor failure. Relation prediction is likewise tied to 3D geometry, through
GNN message passing~\cite{wu2021scenegraphfusion, wu2023incremental}, geometric rules on bounding
boxes or point clouds~\cite{gu2024conceptgraphs, linok2025beyond} or vision-language model (VLM)
captioning of object pairs~\cite{gu2024conceptgraphs}.

We decouple the source of 3D supervision from the inference-time input modality,
confining depth to training, where it supervises a parametric relation predictor
that learns pairwise 3D geometry from 2D visual and geometric cues. At inference
the predictor sees only RGB observations with estimated camera poses and panoptic
masks, producing the measurements that drive pose graph optimization for globally
consistent object positioning, hence preserving the incremental pipeline while removing
the runtime depth dependence.

\subsection{Pose Graph Optimization for Object-Level Mapping}
Pose graph optimization (PGO) formulates spatial estimation as nonlinear least-squares over a factor graph whose nodes are unknown positions and whose 
edges encode pairwise measurements. Developed for camera trajectory estimation in 
SLAM with efficient solvers such as g2o~\cite{kummerle2011g} and 
GTSAM~\cite{dellaert2012factor}, PGO has been extended to object-level mapping, 
where detected objects act as landmarks alongside camera poses. 
QuadricSLAM~\cite{nicholson2018quadricslam} represents objects as ellipsoids 
optimized from bounding box observations, EAO-SLAM~\cite{wu2020eao} folds object 
associations into the pose graph, and CubeSLAM~\cite{yang2019cubeslam} integrates 
monocular 3D cuboid detections into bundle adjustment.

These methods derive their factors from geometric observations such as boxes,
cuboids, depth or known object dimensions. We instead use learned pairwise
predictions, where the relation predictor estimates inter-object distances and angles from
2D images, and PGO optimizes over them, so the factor graph needs no depth sensor.

%% file: sec/3_finalcopy.tex
\section{Methods}

We present IncR-3DSGG, a framework for incremental 3D scene graph 
construction from sequence of posed RGB images. The pipeline operates in two stages: 
per-view\footnote{Each image is treated as an independent view.} local 
scene graph generation and incremental merging into a global scene 
graph representing the current state estimate of the environment. Both, 
local scene graph ($\mathcal{G_L}$) and global scene graph ($\mathcal{G_G}$), are
represented as $\mathcal{G} = (\mathcal{V}, \mathcal{E})$, where 
$\mathcal{V}$ are objects and $\mathcal{E}$ their relationships. Each module is detailed in the following 
subsections. We evaluate on static indoor 
scenes, leaving dynamic environments to future work.

\subsection{Overview}
At each time step $t$ the framework receives an RGB image $\mathcal{I}_t$ with
 camera pose $T_t \in \mathrm{SE}(3)$ and constructs a local scene graph
$\mathcal{G_L} = (\mathcal{V_L}, \mathcal{E_L})$. Object instances are identified
by panoptic segmentation with EoMT~\cite{kerssies2025your}, each forming a node
$v_i \in \mathcal{V_L}$ with its mask and pooled visual feature. For every
co-visible pair, our Parametric Relation Predictor (PRP) predicts an edge
$e_{ij} \in \mathcal{E_L}$ carrying a relation label and the pairwise 3D
geometry in the camera frame. PRP is trained with privileged depth supervision but
requires only posed RGB images at inference.

The global graph $\mathcal{G_G} = (\mathcal{V_G}, \mathcal{E_G})$ is built
incrementally. Detections are matched to existing nodes or inserted as new
instances by Hungarian matching~\cite{kuhn1955hungarian} over visual similarity
with a soft class-consistency constraint. The predicted camera-frame geometry
(centroid distance, azimuth, elevation) is rotated into world-frame displacements
using $T_t$ and used as constraints in a pose graph optimization (PGO) that
jointly recovers globally consistent 3D object positions. Edge attributes are
fused across views by confidence-weighted voting. Figure~\ref{fig:piepile_incr} shows
the complete pipeline.

\subsection{Local Scene Graph Generation}
This section details the construction of the per-view local scene graph 
$\mathcal{G_L}$. The construction focuses on object node detection with EoMT~\cite{kerssies2025your} 
(Sec.~\ref{object_det}) and relation prediction over co-visible object 
pairs with our PRP model (Sec.~\ref{sec:prp}).

\subsubsection{Object Node Detection} \label{object_det}

Object instances in each view are identified by the Encoder-Only Mask Transformer
(EoMT)~\cite{kerssies2025your}, a plain Vision
Transformer~\cite{dosovitskiy2020image} that produces semantic labels, instance
identities and pixel-level masks in a single forward pass at low latency. Unlike
box-based detectors~\cite{robinson2025rf}, its masks enable more precise feature
extraction and more robust cross-view association. We initialize from COCO
panoptic~\cite{kirillov2019panoptic} checkpoints with a
DINOv3~\cite{simeoni2025dinov3} backbone and fine-tune on
3RScan~\cite{wald2019rio} frames using the NYU40
vocabulary~\cite{silberman2012indoor}, of which $39$ classes occur in the dataset
(supplementary B.1). 
Each instance $i$ forms a node
$v_i \in \mathcal{V}_L$ with class $c_i$, mask $m_i$, and feature $f_i$ obtained by masked average pooling of the 
backbone feature map with $m_i$.

\subsubsection{Parametric Relation Predictor}\label{sec:prp}
\begin{figure*}[!ht]
\centering
\includegraphics[width=0.75\textwidth]{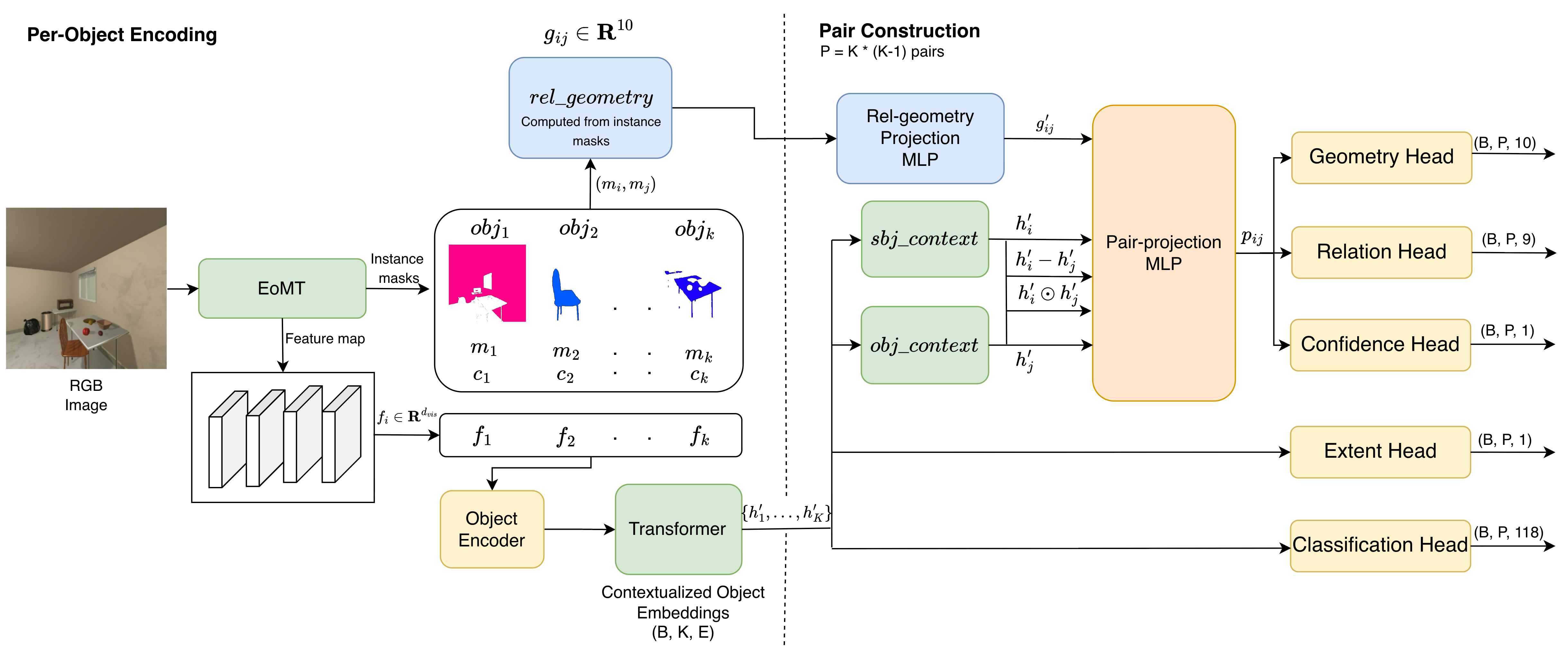}

\caption{Parametric Relation Predictor (PRP) architecture. In the \textit{per-object encoding} stage (left), 
EoMT extracts visual features $f_i$, instance masks $m_i$ and class labels $c_i$ for $K$ detected objects. 
Pairwise 2D geometry $g_{ij}$ is computed from the predicted masks ($m_i$, $m_j$). In the \textit{pair construction} 
stage (right), contextualized subject ($h'_i$) and object ($h'_j$) representations are combined with the 
projected 2D geometry ($g'_{ij}$) to form pair features ($p_{ij}$), which are decoded by independent heads 
predicting 3D geometry, relation class, pairwise confidence, and per-object extent.}
\label{fig:prp}
\end{figure*}

A central challenge in building 3D scene graphs from posed RGB alone, is inferring the pairwise 3D
geometry that relational reasoning depends on with no access to depth. Our
Parametric Relation Predictor (PRP) addresses this by taking 2D visual and
geometric cues and jointly predicting pairwise 3D geometry and a semantic relation
class for every co-visible object pair in the current view.
Figure~\ref{fig:prp} illustrates the architecture, organized into a per-object
stage (left) and a per-pair stage (right).


\noindent\textbf{Input Representation:} As shown on the left of Figure~\ref{fig:prp}, an RGB image is passed through EoMT,
which produces a dense feature map and a set of instance masks. For each of the $K$
detected objects, we extract a mask-pooled visual feature $f_i \in \mathbb{R}^{d_{vis}}$ and its
semantic class label $c_i$. For each
co-visible pair $(i,j)$ we construct a relative geometry vector
$g_{ij} \in \mathbb{R}^{10}$ encoding the pairwise 2D configuration: center
displacement, area ratio, bounding-box IoU, vertical and horizontal displacement,
relative scale, mask overlap, enclosure ratio and 2D centroid distance
(supplementary C.2). These viewpoint-dependent cues are the only
signals from which PRP must recover viewpoint-independent 3D geometry. Targets are
obtained by back-projecting instance masks with depth
(supplementary C.1) and supervise the geometry head
through a Huber loss~\cite{huber1992robust} during training only.

\noindent\textbf{Per-Object Encoding:} Each visual feature $f_i$ is projected to a shared embedding dimension $E$ by a 
two-layer MLP (the \emph{Object Encoder} in Figure~\ref{fig:prp}), and a learned 
category embedding from $c_i$ is added, injecting semantic priors about typical 
spatial configurations (cups rest on tables, books sit inside shelves). The $K$ 
object tokens are then processed by a Transformer encoder in which every object 
attends to every other object in the view. We omit positional encoding here, as the object
tokens form an unordered set. Spatial layout is instead supplied at the pair construction 
stage through relative geometry vector $g_{ij}$. This object-level self-attention is the key context-encoding step, when 
PRP later reasons about a cup, the cup's representation already encodes that a 
table lies below it. The encoder outputs contextualized embeddings 
$\{h'_1, \dots, h'_K\}$ of shape $(B, K, E)$\footnote{$B$: batch size, $K$: object 
detections per view, $E$: embedding dimension.}, shown at the bottom left of 
Figure~\ref{fig:prp}.

\vspace{0.3em}
\noindent\textbf{Pair Construction:}\label{para:pair} The right half of 
Figure~\ref{fig:prp} shows the per-pair stage, which forms $P = K(K{-}1)$ directed 
pairs. For each pair $(i,j)$ the subject and object embeddings are gathered from 
the contextualized tokens and combined with the projected relative geometry to form a pair feature:
\begin{equation}
    p_{ij} = \text{MLP}\big([\,h_i';\; h_j';\; h_i' - h_j';\; h_i' \odot h_j';\; g'_{ij}\,]\big),
\end{equation}
where $[\cdot]$ denotes concatenation and $\odot$ the element-wise product. The 
difference term captures directional displacement in feature space, the product 
term captures feature-level co-activation between the two objects.

\noindent\textbf{Prediction Heads:} Each pair feature $p_{ij}$ is decoded by
several heads (Figure~\ref{fig:prp}, right). The \textbf{Geometry Head} predicts a
10-dimensional vector: centroid geometry (distance $d_c$, azimuth $\alpha_c$,
elevation $\epsilon_c$), which becomes the PGO constraints, and
point-cloud-derived features (closest-point distance, support gap, horizontal gap,
horizontal containment, vertical overlap ratio, enclosure ratio, volume ratio)
capturing contact and containment geometry
(supplementary C.2). The \textbf{Relation Head}
classifies the pair over the eight 3DSSG support predicates including \textit{supported by}, \textit{attached to}, \textit{standing on},
\textit{hanging on}, \textit{connected to}, \textit{part of}, \textit{build in},
\textit{same part} and additionally \textit{none}, trained with inverse-frequency class
weighting and a focal term to counter the dominance of \textit{none}. The
\textbf{Confidence Head} outputs a scalar weighting for each PGO constraint. An
\textbf{Extent Head} on the contextualized embeddings $h'_i$ predicts per-object
spatial extent, so larger objects receive lower constraint weights, and auxiliary
\textbf{classification heads} on $h'_i$ encourage category-discriminative features
in the shared encoder.


\subsubsection{Training Objective}
PRP is trained with a masked multi-task loss combining weighted cross-entropy for
the auxiliary category heads, a focal loss~\cite{lin2017focal} with
inverse-frequency class weighting for the relation head, a Huber
loss~\cite{huber1992robust} over the ten geometric targets and object extents, and
binary cross-entropy for the confidence head. Centroid and closest-point distances
are supervised in log space via $\log(1+d)$ as indoor distances span a wide range,
and the transform prevents large distances from dominating the gradient while
preserving sensitivity to the small distances that determine contact predicates
such as \textit{supported by} and \textit{standing on}. Full loss terms and
weights are given in supplementary D.1.

\subsection{Incremental Scene Graph Construction}
Given the per-view local scene graph $\mathcal{G}_L$, we incrementally construct a global scene graph $\mathcal{G}_G = (\mathcal{V}_G, \mathcal{E}_G)$ that maintains a persistent, metrically grounded representation of the environment. Each new view triggers four operations: object association, node update, position optimization via PGO, and edge aggregation.

\subsubsection{Object Association}
Each local detection is either matched to an existing node in $\mathcal{G}_G$ or
instantiated as a new one. We cast this as a linear assignment problem over a cost
matrix $C \in \mathbb{R}^{N_L \times N_G}$:
\begin{equation}
C_{ij} = w_q\,\bigl(1 - \mathrm{sim}(f_i, \bar f_j)\bigr) + w_c\,\phi(c_i, c_j),
\end{equation}
where $f_i$ is the mask-pooled visual feature of detection $i$, $\bar f_j$ is the running
mean feature of node $j$, and $\mathrm{sim}(\cdot,\cdot)$ is cosine similarity.
The class term $\phi$ is zero for identical classes $c_i = c_j$, a fixed penalty
$\lambda$ for class pairs the segmentation model empirically confuses, and $\infty$
(a hard block) for all other pairs. The assignment is solved with the Hungarian
algorithm~\cite{kuhn1955hungarian}. Matches with $C_{ij} > \tau$ are rejected and
their detections are instantiated as new nodes.

\subsubsection{Node Update}
For matched pairs, the global node feature is updated by exponential moving average:
\begin{equation}
f_j^{(t)} = m \cdot f_j^{(t-1)} + (1 - m) \cdot \hat{f}_i
\end{equation}

\noindent with momentum $m$ and $L_2$-normalized detection feature $\hat{f}_i$. The node label is refined by 
confidence-weighted majority voting over all observations. Unmatched detections are inserted as 
new nodes, initialized from the highest-confidence pairwise constraint linking them to an already-localized node. 
The first node created anchors the PGO coordinate frame.

\subsubsection{Pose Graph Optimization}
PRP provides camera-frame centroid distances, azimuths and elevations for each
co-visible pair of local detections, which the camera pose $T_t$ converts into a
world-frame displacement $\hat{\boldsymbol{\delta}}_{ab} \in \mathbb{R}^3$
between the global nodes $a$ and $b$ they are associated with. Let
$\mathbf{p}_a \in \mathbb{R}^3$ be the position of global node $a$ and
$\mathcal{P} = \{\mathbf{p}_a\}_{a=1}^{N_G}$ the variables to be optimized. Each
pair with an accumulated measurement forms a constraint in $\mathcal{C}$,
weighted by $w_{ab} = \gamma_{ab} / (1 + e_a + e_b)$ with predicted confidence
$\gamma_{ab}$ and extents $e_a, e_b$, so that large objects, whose centroids are
inherently less precise, contribute less. Repeated observations of a pair are
merged by weighted averaging. All positions are then estimated jointly over the residuals
$\mathbf{r}_{ab}(\mathcal{P}) = (\mathbf{p}_b - \mathbf{p}_a)
- \hat{\boldsymbol{\delta}}_{ab}$:
\begin{equation}
\mathcal{P}^{*} = \operatorname*{arg\,min}_{\mathcal{P}}
\sum_{(a,b) \in \mathcal{C}}
\rho_H\!\left( \sqrt{w_{ab}}\, \mathbf{r}_{ab}(\mathcal{P}) \right)
\;+\; w_{\text{a}} \,\bigl\| \mathbf{p}_{a_0} - \mathbf{p}^{0} \bigr\|^2
\label{eq:pgo}
\end{equation}
with $\rho_H$ the component-wise Huber loss. Relative displacements fix
$\mathcal{P}$ only up to a global translation. The second term removes this gauge
freedom by tethering the first observed node $a_0$ to a fixed reference
$\mathbf{p}^{0}$, and is non-constant because $\mathbf{p}_{a_0} \in \mathcal{P}$
is itself optimized. We solve Eq.~\ref{eq:pgo} with trust-region nonlinear least
squares~\cite{branch1999subspace}, re-running after every view.

\subsubsection{Edge Aggregation}
Each directed edge in $\mathcal{G}_G$ accumulates relation predictions across views by
voting weighted with PRP's pairwise confidence, and keeps the label with the highest
cumulative vote. In static scenes, support relations are viewpoint-invariant, so
disagreement across views is predictor noise that voting averages out. Voting is
independent of node positions, whereas PGO affects geometric attributes only. Predicted
closest-point distances and extents are fused by confidence-weighted running means,
while centroid distances and angles are recomputed from the optimized positions after
each PGO step.

%% file: sec/04_results_cam_ready.tex
\section{Results}
We evaluate on two complementary settings. On \textbf{3RScan/3DSSG}~\cite{wald2019rio, wald2020learning} datasets, we compare
against published 3D scene graph methods on real-world scans on the task of 3D scene graph generation.
On \textbf{AI2-THOR}~\cite{kolve2017ai2} we run controlled experiments that real
scans cannot support where ground-truth depth, masks and poses are available at every
frame, allowing us to isolate the merging strategy from input quality and to sweep
viewpoint conditions.

\subsection{Datasets and Experimental Setup}
\noindent\textbf{3RScan/3DSSG:} Our primary evaluation is on
3DSSG~\cite{wald2020learning}, the standard dataset for 3D scene graph estimation,
which augments the real-world RGB-D scans of 3RScan~\cite{wald2019rio} with
per-scene annotations of object instances and their pairwise relations. 
Of the three settings, which differ in whether
instances come from ground-truth mesh segmentation (\textsc{GT}), an RGB-D
reconstruction (\textsc{Dense}), or a point map triangulated from the RGB sequence
by ORB-SLAM3~\cite{campos2021orb} (\textsc{Sparse}), we evaluate under
\textsc{Sparse}, the only one in which every method operates on RGB-derived
geometry and a depth-free method can be compared with prior work on common ground.
We follow its \textsc{l20} configuration on the 157-scan test split, mapping object
classes to the 20 ScanNet classes~\cite{dai2017scannet}, a subset of
NYU40~\cite{silberman2012indoor}, and estimating a single predicate from eight
support relations. 
The baselines read object
instances and geometry from the ORB-SLAM3 point map. We take instances from
panoptic segmentation and use only the ORB-SLAM3 poses, rigidly aligned to the
ground-truth trajectory to share an evaluation frame. 
Depth serves only as training supervision for PRP and
is never used at inference.

\noindent\textbf{AI2-THOR:} We additionally evaluate in the AI2-THOR interactive
simulator~\cite{kolve2017ai2}, spanning four room types (kitchens, living rooms,
bedrooms, bathrooms) with 30 configurations each and split 20/5/5 per room type for
training, validation and inference at roughly 150 images per scene. Exact depth,
masks and poses here support analyses real scans do not, including comparison
against depth-based merging under noise-free input and a viewpoint sweep under
controlled camera rotation.

\noindent\textbf{Training supervision:} PRP is trained with privileged 3D
supervision, where instance masks are projected into 3D using depth and known
intrinsics, and pairwise quantities (centroid distance, azimuth, elevation,
closest-point distance, extent) are computed from the resulting point sets. Depth
is used solely to generate these annotations and is never available at inference.
EoMT is fine-tuned on panoptic annotations derived from the provided instance
masks.

\subsection{3D Scene Graph Generation on 3RScan}
\label{sec:3rscan_sgg}
We compare against five prior methods evaluated in this setting.
IMP~\cite{xu2017scene} computes node features from image regions and edge features
from region unions, updating both by message passing. VGfM~\cite{gay2018visual}
adds geometric features and temporal message passing.
3DSSG~\cite{wald2020learning} replaces image regions with 3D bounding boxes
encoded by PointNet~\cite{qi2017pointnet}, and SGFN~\cite{wu2021scenegraphfusion}
adds a relative-geometry edge descriptor with attention-based aggregation for
incremental estimation. MonoSSG~\cite{wu2023incremental} fuses multi-view image
features with geometry from an incrementally reconstructed sparse point map.
 
Every baseline reads object centroids, extents and pairwise offsets from
the triangulated ORB-SLAM3 point map. In our method, PRP predicts these quantities from image evidence and PGO
fuses them into globally consistent positions. Instances come from panoptic segmentation,
which the \textsc{Sparse} setting permits since it constrains the input modality
rather than the detector. 

\begin{table}[t]
\centering
\resizebox{\columnwidth}{!}{%
\begin{tabular}{lccccccc}
\toprule
 & \multicolumn{3}{c}{Recall (R@1)} & \multicolumn{2}{c}{Annotated} & \multicolumn{2}{c}{Mean Recall} \\
\cmidrule(lr){2-4} \cmidrule(lr){5-6} \cmidrule(lr){7-8}
Method & Obj & Pred & Trip & Pred$^*$ & Trip$^*$ & mR.Obj & mR.Pred \\
\midrule
IMP~\cite{xu2017scene}             & 27.5 & \textbf{90.8} & 8.5 & 4.5 & 3.9 & 20.4 & 14.0 \\
VGfM~\cite{gay2018visual}          & 27.7 & 90.6 & 8.8 & 4.3 & 3.9 & 24.5 & 13.9 \\
3DSSG~\cite{wald2020learning}      & 10.8 & 90.1 & 1.3 & 4.1 & 1.0 & 6.0  & 13.7 \\
SGFN~\cite{wu2021scenegraphfusion} & 15.4 & 88.3 & 2.4 & 7.0 & 3.3 & 8.6  & 15.5 \\
MonoSSG~\cite{wu2023incremental}   & 28.6 & 89.8 & 9.8 & 8.4 & 6.9 & 27.0 & 17.6 \\
\midrule
\textbf{Ours}                      & \textbf{39.5} & 88.6 & \textbf{14.5} & \textbf{28.3} & \textbf{23.1} & \textbf{52.2} & \textbf{30.8} \\
\bottomrule
\end{tabular}%
}
\caption{3D scene graph generation on 3RScan (\textsc{Sparse}, \textsc{l20}, 157 test
scans, official sets). Baselines read geometry from the ORB-SLAM3 point map, whereas ours uses
no depth or reconstructed geometry at inference. Instances come from each method's own
front end (ours: EoMT). Pred$^*$, Trip$^*$ scores annotated pairs only.}
\label{tab:3dssg_sparse}
\end{table}

\noindent\textbf{Protocol:}
Following prior
work~\cite{wald2020learning,wu2021scenegraphfusion,wu2023incremental}, we report
Recall@1 for objects (Obj), predicates (Pred) and triplets (Trip), counting a
triplet correct only when subject, object and predicate all match. We score the
official sets of 2{,}316 objects and 20{,}904 neighbour-based pairs, so every
method shares the same denominator. An undetected object receives a \texttt{none}
prediction and counts as an error against its ground-truth class, and an
unpredicted pair is likewise assigned \texttt{none}.
Since most pairs carry no relation, Pred and Trip are dominated by
\texttt{none}. We therefore also report Pred$^*$ and Trip$^*$, computed only on pairs with an annotated relation, excluding pairs labelled
\texttt{none}. Mean
recall (\textit{mR.Obj, mR.Pred}) averages per-class recall over classes with
support and includes \texttt{none} for predicates, following the official code.

\noindent\textbf{Object recovery:}
As shown in Table \ref{tab:3dssg_sparse}, our method reaches $39.5$ object recall against $28.6$ for MonoSSG, and the margin
widens on mean object recall ($52.2$ vs.\ $27.0$), which weights infrequent
classes. Per-class recall rises most on sink ($86.7$ vs.\ $18.3$), sofa ($73.7$
vs.\ $23.7$) and toilet ($74.1$ vs.\ $29.6$). Walls are the exception ($11.9$ vs.\
$32.5$), as large planar segments are often split or merged across views. The baselines separate along the instance
source, where point-based methods (3DSSG $10.8$, SGFN $15.4$) fall far behind
image-based ones ($27.5$--$28.6$). What the instance source captures therefore
matters more for object recovery than how relations are modelled, and part of our
margin follows from recovering instances that the sparse map never represents.
 
\noindent\textbf{Predicate and triplet estimation:}
Triplet recall is $14.5$, above MonoSSG's $9.8$ and well above the point-based
methods, and mean predicate recall nearly doubles ($30.8$ vs.\ $17.6$). The
\texttt{none}-dominated columns understate this difference. On annotated pairs we
reach $28.3$ predicate and $23.1$ triplet recall, against at most $8.4$ and $6.9$
for the baselines, which predict \texttt{none} for most annotated pairs (see Table \ref{tab:3dssg_sparse}). 
Most of this gain persists when PRP's geometry is replaced by the sparse map
($22.1$ Pred$^*$, Table~\ref{tab:ablation_sources}), so it stems from the panoptic
front end together with PRP's relation head, while PRP's geometry adds a further
$6.2$ points by covering pairs the map cannot triangulate.
 
\begin{table}[t]
\centering
\footnotesize
\resizebox{\columnwidth}{!}{%
\begin{tabular}{llcccccccc}
\toprule
 & & \multicolumn{3}{c}{Recall (R@1)} & \multicolumn{2}{c}{Annotated} & \multicolumn{2}{c}{Mean Recall} & \\
\cmidrule(lr){3-5} \cmidrule(lr){6-7} \cmidrule(lr){8-9}
Instances & Geometry & Obj & Pred & Trip & Pred$^*$ & Trip$^*$ & mR.Obj & mR.Pred & PosErr (m) \\
\midrule
EoMT & sparse map & 39.5 & 89.7 & 15.3 & 22.1 & 18.0 & 52.2 & 27.0 & 1.286 \\
EoMT & PRP        & 39.5 & 88.6 & 14.5 & \textbf{28.3} & \textbf{23.1} & 52.2 & \textbf{30.8} & \textbf{1.242} \\
\midrule
GT   & sparse map & 67.1 & 86.7 & 38.7 & 36.0 & 36.0 & 84.9 & 42.4 & -- \\
GT   & PRP        & 68.4 & 82.8 & 36.7 & 52.1 & 52.1 & 85.3 & 55.3 & -- \\
\bottomrule
\end{tabular}%
}
\caption{Instance source vs.\ geometry source on 3RScan, with all other components
fixed. \emph{Sparse map} replaces PRP's pairwise geometry with the ORB-SLAM3 point
map. Position error is reported for predicted instances only.}
\label{tab:ablation_sources}
\end{table}
 
\noindent\textbf{Instance and geometry source:}
Table~\ref{tab:ablation_sources} varies the instance source and the geometry
source with all other components fixed. Replacing PRP geometry with the sparse map
leaves every object metric unchanged, since association and class voting do not
use relation geometry. Our object gain over the baselines therefore comes from the
instance front end and association, not from PRP. Geometry instead affects
relations and localization. With EoMT instances, PRP beats the sparse map by $6.2$
points on Pred$^*$ and $5.1$ on Trip$^*$, since the map places only 69.6\% of
detected instances while PRP covers every co-visible pair regardless of whether
its surfaces triangulate. PRP also lowers position error slightly, from $1.29$ to $1.24$\,m
(Table~\ref{tab:ablation_sources}). The sparse variant scores slightly higher on Trip
($15.3$ vs.\ $14.5$) and Pred ($89.7$ vs.\ $88.6$) only because the pairs it cannot
place become \texttt{none}, which matches most ground truth.
 
Substituting ground-truth masks and classes, with PRP and PGO unchanged, raises
object recall to $68.4$, mean object recall to $85.3$ and triplet recall to
$36.7$. Annotated-pair recall nearly doubles ($28.3 \rightarrow 52.1$), and
Trip$^*$ equals Pred$^*$, since every recovered object is now correctly classified
and a triplet fails only through its predicate. Pred drops to $82.8$ because more
recovered objects produce more predicted edges on \texttt{none} pairs. With the
same instances the sparse map reaches only $36.0$ on both, so PRP's advantage widens to $16.1$
points.  The relation
model thus performs well once its endpoints are correct, and front-end quality,
not relation prediction, is the main limit on triplet recall.

\begin{figure}[t]
    \centering
    \includegraphics[width=0.5\columnwidth]{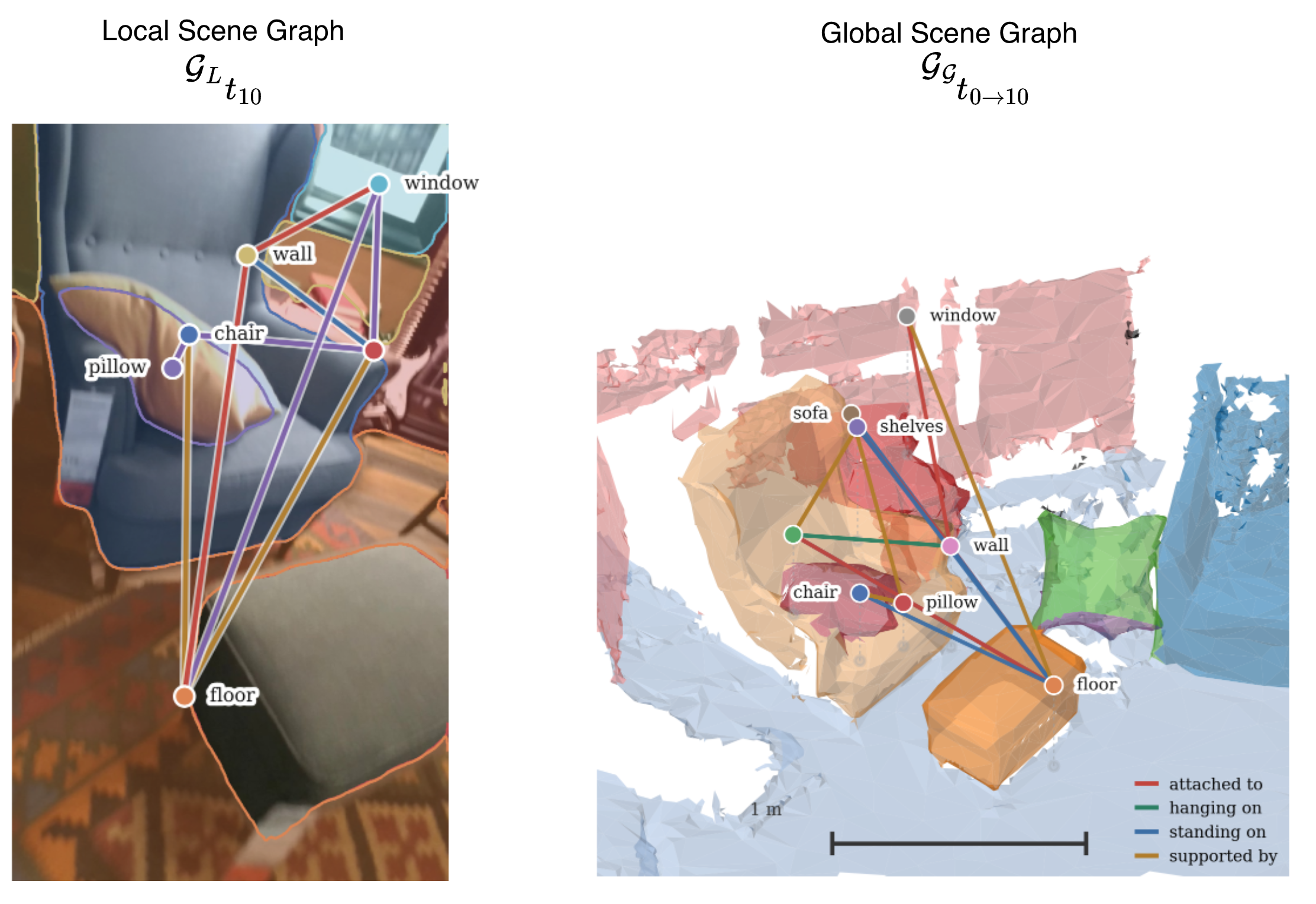}
    \caption{Qualitative example on 3RScan. \textbf{Left:} local scene graph
$\mathcal{G}_L$ at frame $t{=}10$. \textbf{Right:} global scene graph
$\mathcal{G}_G$ over frames $0$--$10$, with nodes at PGO-optimized positions and
edge colour denoting the predicted relation. The mesh is for reference only
(more examples in supplementary G.}
\label{fig:local_global}
\end{figure}

\subsection{Object Localization and Association}
\label{sec:localization}

\noindent\textbf{Baselines:} To isolate association from detection, we take only
the association rule of ConceptGraphs~\cite{gu2024conceptgraphs} (point overlap
with feature similarity) and SGFN~\cite{wu2021scenegraphfusion} (3D proximity) and
run both on our EoMT detections with depth back-projection. These \emph{style}
baselines read depth at every frame, while ours reads none.

\noindent\textbf{Metrics:} \textit{GT recall} is the fraction of annotated objects
recovered as a node. \textit{Node purity} is the mean share of a node's
observations that belong to its dominant ground-truth object. \textit{Fragmentation} is the
mean number of nodes per recovered object ($1.0$ is ideal). \textit{Position
error} is the mean distance to the full-object ground-truth centroid after a
per-scan similarity alignment~\cite{schonemann1966generalized}. PRP is trained on
visible-surface centroids, so part of this error reflects the offset between the
two. Further details on how each metric is computed are given in
supplementary E.1.

\begin{table}[t]
\centering
\footnotesize
\resizebox{\columnwidth}{!}{%
\begin{tabular}{@{}lccccccccc@{}}
\toprule
& & \multicolumn{4}{c}{3RScan~\cite{wald2019rio}} & \multicolumn{3}{c}{AI2-THOR~\cite{kolve2017ai2}} \\
\cmidrule(lr){3-6} \cmidrule(l){7-9}
Method & Depth & GTRec$\uparrow$ & NodePur$\uparrow$ & Frag.$\downarrow$ & PosErr$\downarrow$ & NodePur$\uparrow$ & Frag.$\downarrow$ & PosErr$\downarrow$ \\
\midrule
ConceptGraphs-style~\cite{gu2024conceptgraphs} & \checkmark & 0.590 & \textbf{0.971} & 1.736 & \textbf{0.719} & \textbf{0.978} & 1.196 & 0.646 \\
SGFusion-style~\cite{wu2021scenegraphfusion}   & \checkmark & \textbf{0.710} & 0.929 & 1.717 & 0.764 & 0.964 & 1.758 & \textbf{0.591} \\
\textbf{Ours}                                  & -- & 0.500 & 0.813 & \textbf{1.155} & 1.242 & 0.936 & \textbf{1.116} & 0.692 \\
\bottomrule
\end{tabular}%
}
\caption{Object localization and association. Baselines use our EoMT masks with
depth back-projection. Ours uses posed RGB only. Position error in metres.}
\label{tab:localization}
\end{table}

\noindent\textbf{3RScan:}
The depth-based baselines recover more objects ($0.590$--$0.710$ vs.\ $0.500$),
have higher node purity ($0.929$--$0.971$ vs.\ $0.813$) and localize more precisely
($0.719$--$0.764$\,m vs.\ $1.242$\,m), as back-projected depth measures position
directly. Our method produces fewer fragments per recovered object ($1.155$ vs.\
$1.717$--$1.736$). As fragmentation counts recovered objects only, this is not an
overall association advantage. With detections fixed, PRP geometry localizes
slightly better than the sparse map ($1.242$ vs.\ $1.286$\,m,
Table~\ref{tab:ablation_sources}), a small gain over the triangulated points
available without depth.

\noindent\textbf{Robustness:}
On AI2-THOR~\cite{kolve2017ai2} with predicted masks, the depth-based baselines
localize as well as or slightly better than ours ($0.591$--$0.646$ vs.\ $0.692$\,m)
with higher node purity ($0.964$--$0.978$ vs.\ $0.936$), while ours fragments
objects least ($1.116$ vs.\ $1.196$--$1.758$). With ground-truth masks the
baselines localize better (supplementary E.2). Under random camera
yaw, however, their node count grows from $35$ to $929$, against $29$ to $54$ for
ours (supplementary E.3), as point-cloud overlap between views
breaks down. Our advantage is thus robust association without depth, not more
accurate positions.

\subsection{Runtime and Additional Analyses}

\noindent\textbf{Runtime:} On a single RTX 3090, a view takes $1024$\,ms on average
over the 3RScan split, of which panoptic segmentation accounts for $1008$\,ms. PRP
($2.3$\,ms), association ($1.4$\,ms) and PGO ($12.5$\,ms) add under $2\%$ together,
so the relational components cost little on top of a segmentation front end an
embodied system would run regardless.

\noindent\textbf{Additional analyses:} In the supplementary, PRP's architectural
ablations show that removing object self-attention or simplifying pair construction
degrades every metric only slightly (supplementary D.2). PRP is
also compared against VGGT~\cite{wang2025vggt}, whose point maps are not metric and
need a per-scan scale fitted from ground truth that PRP never needs. Even with that
correction, PRP is more accurate on $124$ of $155$ scans
(supplementary F).

%% file: sec/5_conclusion.tex
\section{Conclusion}
We presented IncR-3DSGG, an incremental 3D scene graph generation pipeline
combining panoptic segmentation, a learned parametric relation predictor and pose
graph optimization to build metrically grounded scene graphs from posed RGB alone,
without depth at inference. On the 3DSSG \textsc{Sparse} benchmark it exceeds prior
methods on object recall ($39.5$ vs.\ $28.6$), triplet recall ($14.5$ vs.\ $9.8$)
and mean predicate recall ($30.8$ vs.\ $17.6$). Ablations attribute the object gain
to the panoptic front end and the relational gain to the relation predictor, which
beats a sparse point map on annotated relations ($28.3$ vs.\ $22.1$ Pred$^*$).
Relations can thus be predicted from 2D observations, although depth-based
association still localizes objects more precisely on 3RScan.

Instance coverage remains the binding constraint, since triplet recall is bounded
by how many annotated objects are detected and associated, and the system
relies on accurate camera poses and about one second per keyframe.
Coupling the pipeline with an exploration policy that selects views to reduce
uncertainty in the current graph is a natural next step toward embodied use.